\documentclass[10pt,twocolumn,letterpaper]{article}

\usepackage{cvpr}              

\usepackage{graphicx}
\usepackage{amsmath}
\usepackage{amssymb}
\usepackage{booktabs}
\usepackage{adjustbox}
\usepackage{chngcntr}
\usepackage{multirow}
\usepackage{makecell}
\usepackage[accsupp]{axessibility}
\usepackage{xcolor}
\definecolor{citecolor}{HTML}{2980b9}
\definecolor{linkcolor}{HTML}{c0392b}
\definecolor{dorange}{HTML}{ff6103}
\makeatletter
  \newcommand\figcaption{\def\@captype{figure}\caption}
  \newcommand\tabcaption{\def\@captype{table}\caption}
\makeatother
\usepackage[pagebackref=true,breaklinks=true,colorlinks,bookmarks=false,citecolor=citecolor,linkcolor=linkcolor]{hyperref}
\usepackage{xcolor}         
\usepackage{color, colortbl}
\usepackage{float}

\usepackage[capitalize]{cleveref}
\crefname{section}{Sec.}{Secs.}
\Crefname{section}{Section}{Sections}
\Crefname{table}{Table}{Tables}
\crefname{table}{Tab.}{Tabs.}

\def\cvprPaperID{*****} 
\def\confName{CVPR}
\def\confYear{2023}

\begin{document}

\title{Point-Bind \& Point-LLM: Aligning Point Cloud with Multi-modality for\\3D Understanding, Generation, and Instruction Following}

\author{Ziyu Guo$^{*1}$, Renrui Zhang$^{*\dagger\ddagger1,2}$, Xiangyang Zhu$^{2}$, Yiwen Tang$^{2}$, Xianzheng Ma$^{2}$, Jiaming Han$^{1,2}$\\Kexin Chen$^{1}$, Peng Gao$^{2}$, Xianzhi Li$^{\ddagger3}$, Hongsheng Li$^{1}$, Pheng-Ann Heng$^{1}$\vspace{0.2cm}\\
\normalsize{$^*$ Equal contribution}\quad  $\dagger$ Project leader\quad  $\ddagger$ Corresponding author\vspace{0.3cm}\\
  $^1$The Chinese University of Hong Kong\quad
  $^2$Shanghai AI Laboratory\\
  $^3$Huazhong University of Science and Technology\vspace{0.2cm}\\
\texttt{\{zyguo, pheng\}@cse.cuhk.edu.hk,}\quad \texttt{zhangrenrui@pjlab.org.cn}
}
\maketitle

\begin{abstract}
We introduce \textbf{Point-Bind}, a 3D multi-modality model aligning point clouds with 2D image, language, audio, and video.
Guided by ImageBind, we construct a joint embedding space between 3D and multi-modalities, enabling many promising applications, e.g., any-to-3D generation, 3D embedding arithmetic, and 3D open-world understanding.
On top of this, we further present \textbf{Point-LLM}, \textit{\textbf{the first}} 3D large language model (LLM) following 3D multi-modal instructions.
By parameter-efficient fine-tuning techniques, Point-LLM injects the semantics of Point-Bind into pre-trained LLMs, e.g., LLaMA, which \textbf{requires no 3D instruction data}, but exhibits superior 3D and multi-modal question-answering capacity.
We hope our work may cast a light on the community for extending 3D point clouds to multi-modality applications. Code is available at \url{https://github.com/ZiyuGuo99/Point-Bind_Point-LLM}.

\end{abstract}

\section{Introduction}
In these years, 3D vision has gained significant attention and development, driven by the rising popularity of autonomous driving~\cite{chen20203d,shao2023safety,shi2020pv}, navigation~\cite{tan2001exploring,wang2019reinforced,zhu2020vision}, 3D scene understanding~\cite{armeni20163d,liu2021group,misra2021end,vu2022softgroup}, and robotics~\cite{huang2023voxposer,savva2019habitat}. 
To extend its application scenarios, numerous efforts~\cite{zhang2023meta,guo2023joint,zhang2023learning,afham2022crosspoint} have been made to incorporate 3D point clouds with data from other modalities, allowing for improved 3D understanding~\cite{guo2023joint, afham2022crosspoint}, text-to-3D generation~\cite{nichol2022point,poole2022dreamfusion,lin2023magic3d}, and 3D question answering~\cite{azuma2022scanqa, Hong_2023_CVPR}. 

\begin{figure}[t!]
    \vspace{0.1cm}
    \centering
    \includegraphics[width=\linewidth]{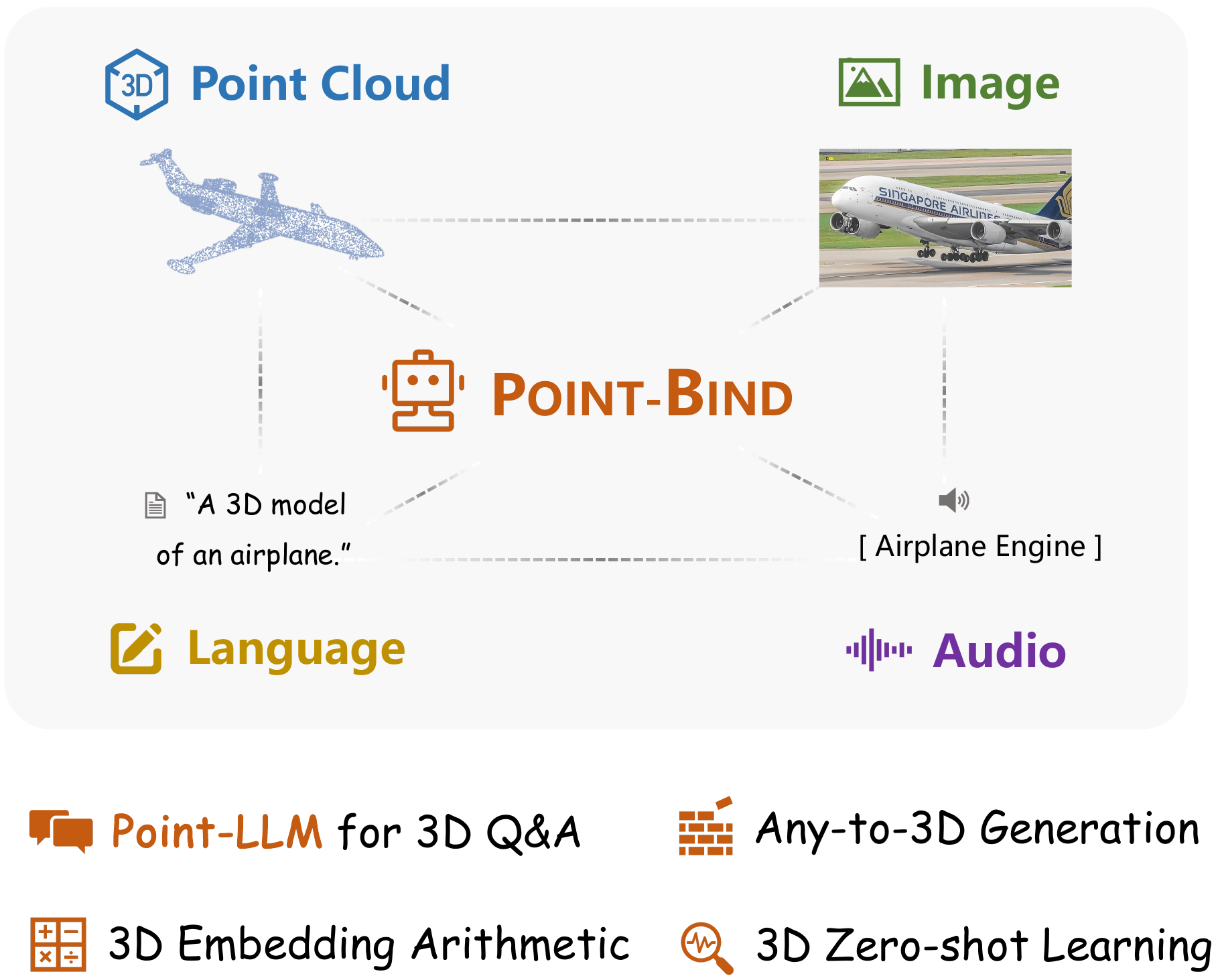}
    \vspace{-0.2cm}
    \caption{\textbf{Characteristics of Point-Bind.} We propose to align 3D with multi-modalities and develop a unified framework, Point-Bind, which extends various 3D multi-modal applications. Based on Point-Bind, we further introduce Point-LLM, a 3D large language model with bilingual 3D instruction-following capacity.}
    \label{teaser}
\end{figure}

\begin{figure*}[t!]
\centering
\includegraphics[width=\textwidth]{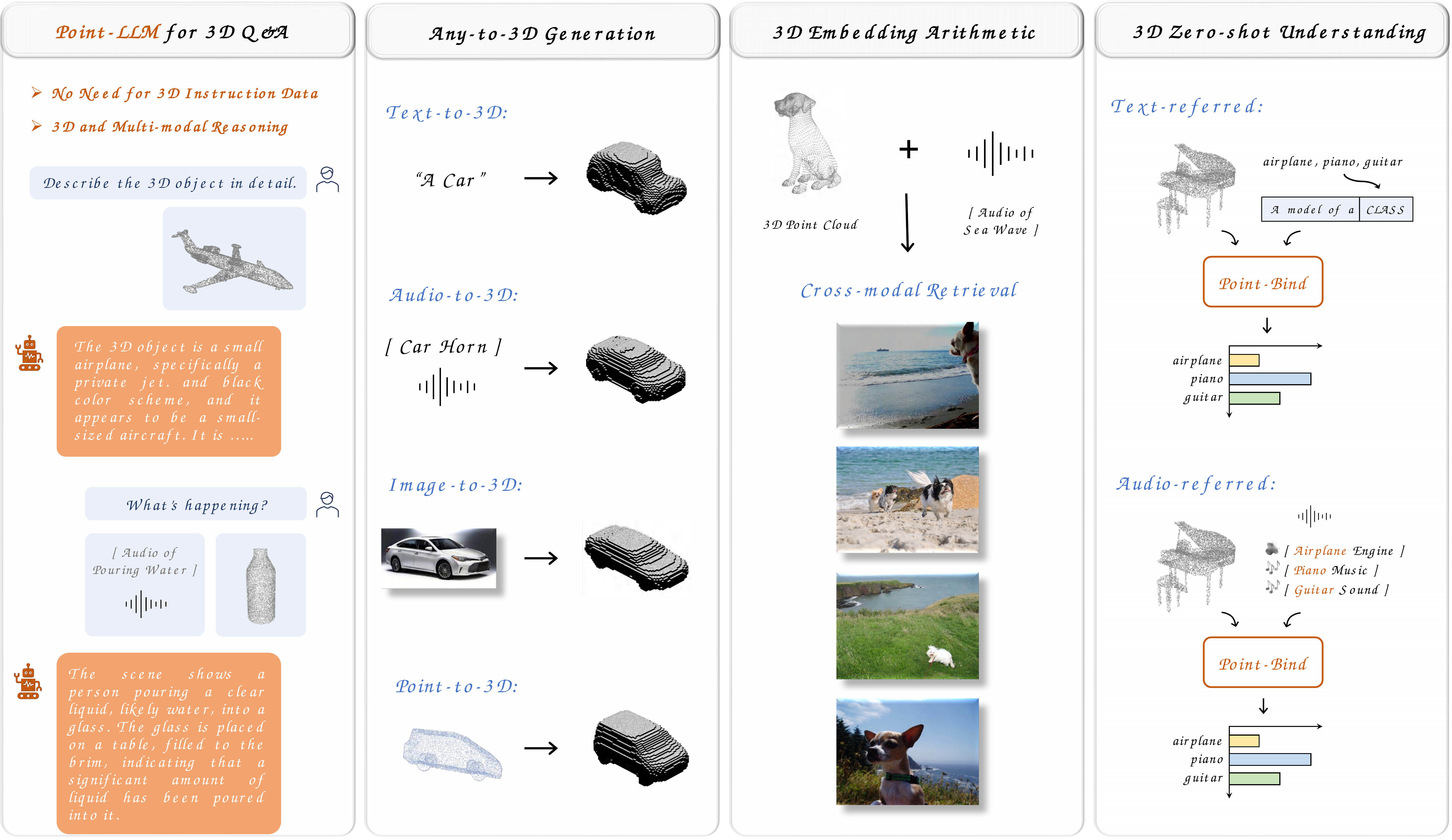}
    \vspace{0.1cm}
   \figcaption{\textbf{3D Multi-modal Applications of Point-Bind.} With a joint 3D multi-modal embedding space, Point-Bind enables many promising application scenarios, e.g., Point-LLM for 3D instruction following, 3D generation conditioned on any modalities, embedding-space arithmetic with 3D, and multi-modal 3D zero-shot understanding.}
    \label{intro}
    \vspace{0.2cm}
\end{figure*}

For 3D geometry understanding, previous works either leverage 2D-language embeddings to guide 3D open-world recognition~\cite{zhang2022pointclip,Zhu2022PointCLIPV2}, or harness visual and textual semantics to assist 3D representation learning~\cite{qi2023recon, xue2022ulip, liu2023openshape}. 
However, their perception capabilities are mostly constrained by limited modalities provided in the training phase.
Inspired by 2D generative models~\cite{ramesh2022hierarchical,saharia2022photorealistic,rombach2022high}, a collection of methods~\cite{poole2022dreamfusion,lin2023magic3d,nichol2022point} has achieved text-to-3D synthesis with high quality and efficiency. Despite this, they lack the ability to generate 3D shapes conditioned on multi-modal input, i.e., any-to-3D generation.
Another series of works connects descriptive natural language with 3D data, which is applied to 3D captioning~\cite{yuan2022x,chen2023end}, question answering~\cite{wijmans2019embodied, azuma2022scanqa}, and visual grounding~\cite{guo2023viewrefer,wu2023eda}. Yet, they fail to utilize the pre-trained linguistic knowledge within large language models (LLMs) to better capture 3D geometrics.


Therefore, how to develop a unified 3D framework aligning with multi-modality for general 3D learning still remains an open question. Very recently, ImageBind~\cite{girdhar2023imagebind} was proposed to learn a shared representation space across six different modalities, i.e., images, text, audio, depth, thermal, and IMU data. 
Motivated by this, we ask the following question: \textbf{\textit{can we construct a joint embedding space between 3D and multi-modality for unified 3D understanding, generation, and insturction following?}}

\vspace{0.1cm}
To this end, we introduce \textbf{Point-Bind}, a 3D multi-modality framework that aligns point clouds with multiple modalities for general 3D analysis, as shown in Figure~\ref{teaser}. Specifically, we collect 3D-image-text-audio pairs as the training data, and construct a joint embedding space guided by ImageBind.
We adopt a contrastive loss between the extracted features from a trainable 3D encoder, e.g., I2P-MAE~\cite{zhang2023learning}, and the frozen multi-modal encoders of ImageBind. Such a simple strategy can efficiently integrate different modalities into a unified representation space, and allows for various 3D-centric multi-modal tasks in Figure~\ref{intro}. 

The main contributions of Point-Bind are as follows:

\begin{itemize}
    \item \textcolor{dorange}{\textbf{\textit{Aligning 3D with ImageBind.}}} Within a joint embedding space, Point-Bind firstly aligns 3D point clouds with multi-modalities guided by ImageBind, including 2D images, video, language, audio, etc.
    \item \textcolor{dorange}{\textbf{\textit{Any-to-3D Generation.}}} Based on existing text-to-3D generative models, Point-Bind enables 3D shape synthesis conditioned on any modalities, i.e., text/image/audio/point-to-mesh generation.
    \item \textcolor{dorange}{\textbf{\textit{3D Embedding-space Arithmetic.}}} We observe that 3D features from Point-Bind can be added with other modalities to incorporate their semantics, achieving composed cross-modal retrieval.
    \item \textcolor{dorange}{\textbf{\textit{3D Zero-shot Understanding.}}} Point-Bind attains \textit{state-of-the-art} performance for 3D zero-shot classification. Also, our approach supports audio-referred 3D open-world understanding, besides text reference.
\end{itemize}

\begin{figure*}[t!]
\centering
\vspace{-0.2cm}
\includegraphics[width=\textwidth]{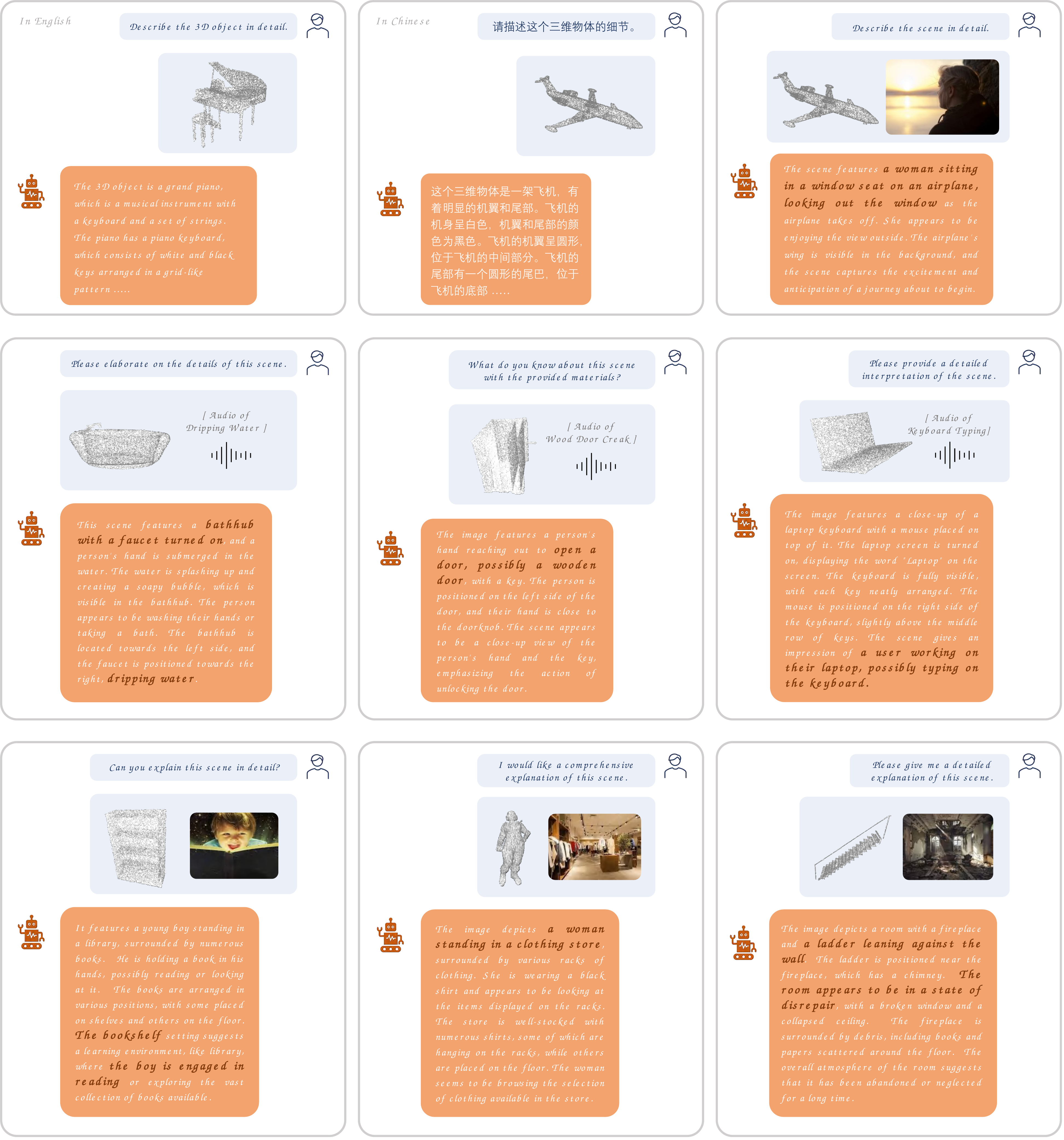}
   \figcaption{\textbf{3D Question-answering Examples of Point-LLM.} Given 3D and multi-modal instructions, our Point-LLM can effectively generate detailed responses and conduct superior cross-modal reasoning. Notably, we do not need any 3D instruction data for training.}
    \label{3dllm}
\end{figure*}

Furthermore, on top of our joint embedding space, we propose to incorporate Point-Bind with LLaMA~\cite{touvron2023llama} to develop \textit{\textbf{the first}} 3D large language models (LLMs), termed as \textbf{Point-LLM}. As shown in Figure~\ref{intro}, our Point-LLM can respond to language instructions with 3D point cloud conditions, and effectively capture spatial geometry characteristics. Referring to ImageBind-LLM~\cite{ImageBind-LLM2023}, we utilize a bind network along with a visual cache model to bridge Point-Bind with LLaMA, and adopt zero-initialized gating mechanisms~\cite{zhang2023llama,gao2023llama} for parameter-efficient fine-tuning. With superior data efficiency, the entire training phase of Point-LLM \textbf{\textit{requires no 3D instruction dataset}}, and only utilizes public vision-language data~\cite{chen2015microsoft,changpinyo2021conceptual, sharma2018conceptual, schuhmann2022laion} for vision-language tuning. In this way, we enable LLMs to understand and conduct cross-modal reasoning for 3D and multi-modal data, achieving superior 3D question-answering capacity in both English and Chinese. 

The main contributions of Point-LLM are as follows:
\begin{itemize}
    \item \textcolor{dorange}{\textbf{\textit{Point-LLM for 3D Question Answering.}}} Using Point-Bind, we introduce Point-LLM, the first 3D LLM that responds to instructions with 3D point cloud conditions, supporting both English and Chinese.
    \item \textcolor{dorange}{\textbf{\textit{Data- and Parameter-efficiency.}}} We only utilize public vision-language data for tuning without any 3D instruction data, and adopt parameter-efficient fine-tuning techniques, saving extensive resources.
    \item \textcolor{dorange}{\textbf{\textit{3D and Multi-modal Reasoning.}}} Via the joint embedding space, Point-LLM can generate descriptive responses by reasoning a combination of 3D and multi-modal input, e.g., a point cloud with an image/audio.
\end{itemize}

\vspace{0.1cm}
\section{Related Work}
\vspace{0.1cm}

\paragraph{Multi-modality Learning.}
Compared to single-modal approaches, multi-modal learning aims to learn from multiple modalities simultaneously, achieving more robust and diverse representation learning. Numerous studies have proved its efficacy, involving 2D images, videos, texts, and audio~\cite{desai2021virtex, fang2021clip2video, nagrani2022learning}, and enhance the cross-modal performance for downstream tasks~\cite{lin2021exploring,Ramesh2021Vset,botach2022endtoend,guo2023calip}, and video-text-audio integration for text generation~\cite{lin2021vx2text}. The representative vision-language pre-training, CLIP~\cite{radford2021learning}, effectively bridges the gap between 2D images and texts, which encourages further exploration of cross-modality learning. Recently, ImageBind~\cite{girdhar2023imagebind} successfully aligns six modalities in a joint embedding space, unleashing the power for emergent zero-shot cross-modal capabilities. However, ImageBind fails to investigate its efficacy on 3D point clouds. In the 3D domain, most existing cross-modal works introduce vision-language alignment~\cite{zhang2022pointclip,xue2022ulip,afham2022crosspoint,guo2023joint,chen2023pimae} into 3D point clouds, and mainly focus on open-world recognition tasks, which ignore the potential of multi-modal semantics for wider 3D applications.
In this paper, our Point-Bind develops a general 3D multi-modality model that aligns 3D point clouds with six other modalities guided by ImageBind, allowing for more diverse 3D cross-modal understanding.

\paragraph{Large Models in 3D.} Large-scale pre-trained models have achieved remarkable downstream performance in language and 2D image processing. Inspired by this, many efforts have introduced 2D and language large models, to assist in 3D learning.
The prior PointCLIP series~\cite{zhang2022pointclip,Zhu2022PointCLIPV2,huang2022clip2point} project 3D point clouds into depth maps, and utilize CLIP~\cite{radford2021learning} for zero-shot recognition.
Image2Point~\cite{xu2022image2point} instead converts 2D pre-trained models into 3D space as a good network initialization.
By contrastive learning, ULIP series~\cite{xue2022ulip,xue2023ulip2} and other works~\cite{liu2023openshape,hegde2023clip} pre-train 3D networks guided by the vision-language embedding space of CLIP.
Another branch of work employs CLIP to guide the text-conditioned generation of 3D objects~\cite{jain2021dreamfields, sanghi2021clip, xu2023dream3d, liu2023iss++} or stylized meshes \cite{mohammad2022clip, text2mesh} by encoding descriptive textual input. 
Some works also adopt GPT-3~\cite{NEURIPS2020_1457c0d6
} to enhance the language-based understanding of 3D spatial geometry, such as PointCLIP V2~\cite{Zhu2022PointCLIPV2} and ViewRefer~\cite{guo2023viewrefer}.
Different from them, we utilize ImageBind~\cite{girdhar2023imagebind} to construct a joint embedding space between 3D point clouds and multiple modalities. The derived Point-Bind can well leverage the multi-modal semantics for general 3D cross-modal understanding, generation, and question answering.

\begin{figure*}[t!]
\centering
\includegraphics[width=\textwidth]{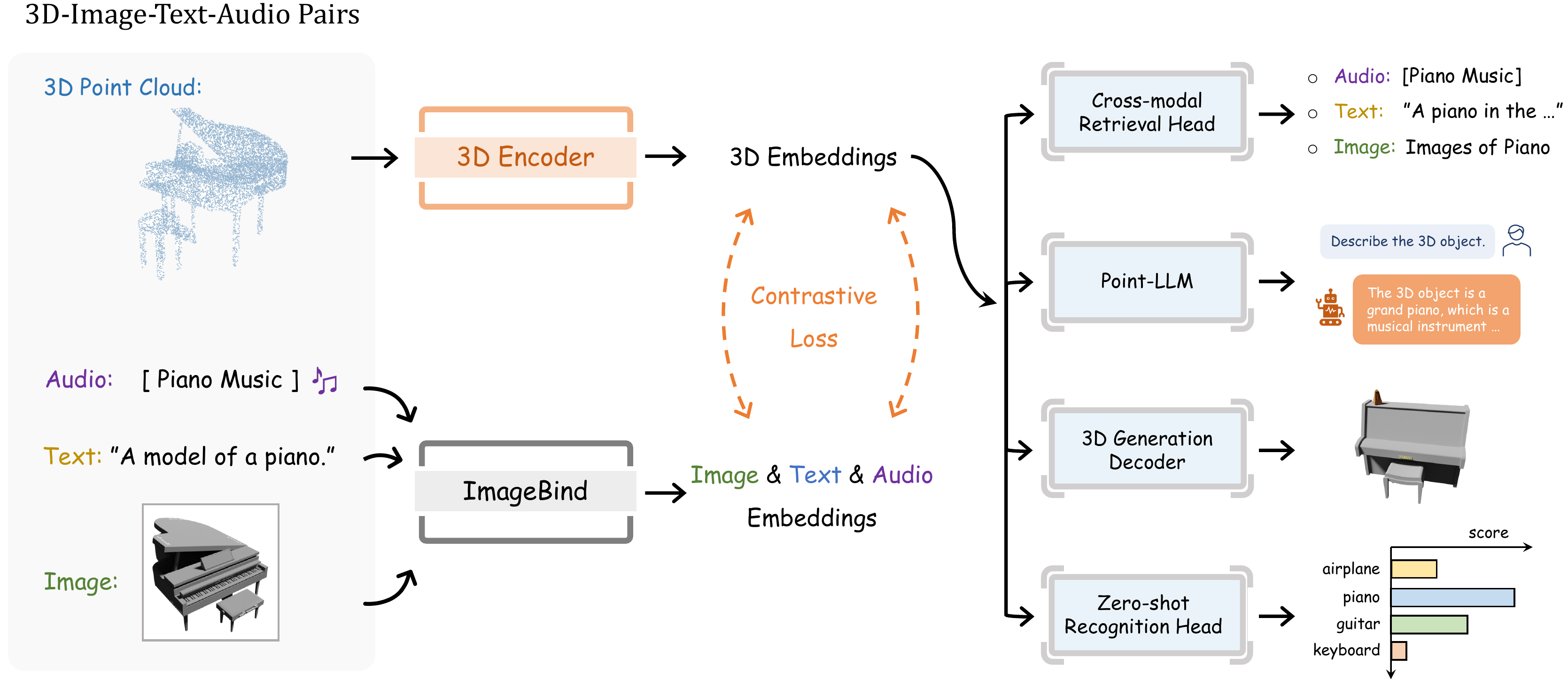}
   \figcaption{\textbf{Overall Pipeline of Point-Bind.} We collect 3D-image-audio-text data pairs for contrastive learning, which aligns 3D modality with others guided ImageBind~\cite{girdhar2023imagebind}. With a joint embedding space, Point-Bind can be utilized for 3D cross-modal retrieval, any-to-3D generation, 3D zero-shot understanding, and developing a 3D large language model, Point-LLM.}
    \label{pipeline}
    \vspace{0.2cm}
\end{figure*}

\paragraph{Pre-training in 3D.}
In recent years, significant progress has been made in supervised learning for 3D vision tasks~\cite{qi2016pointnet, qi2017pointnetplusplus, qian2022pointnext,zhang2023parameter,zhu2023less}. However, these approaches lack satisfactory generalization capabilities for out-of-domain data. 
To address this, self-supervised learning has emerged as a promising solution to enhance 3D transfer learning~\cite{chen2023pimae, yu2021pointbert, li2019pugan, poursaeed2020self}. Most self-supervised pre-training methods employ an encoder-decoder framework to encode point clouds into latent representations and then reconstruct the original data form~\cite{sauder2019selfsupervised, OcCo, rao2020global}. Therein, Point-MAE~\cite{pang2022masked} and Point-M2AE~\cite{zhang2022point} introduce masked autoencoders~\cite{MaskedAutoencoders2021} into 3D point clouds pre-training, achieving competitive results on different 3D tasks.
Alternatively, cross-modal pre-training approaches are also leveraged to enhance the 3D generalization ability~\cite{wang2022p2p, qian2022pix4point, liu2021learning, qi2023recon}. For example, ACT~\cite{dong2022autoencoders} and I2P-MAE~\cite{zhang2023learning} utilize pre-trained 2D transformers as teachers to guide 3D representation learning.
Inspired by previous works, we adopt collected 3D-image-text-audio pairs for self-supervised pre-training, and regard ImageBind's encoders as guidance for contrastive learning. In this way, the Point-Bind is pre-trained to obtain a joint embedding space between 3D and multi-modality, allowing for superior performance on different 3D downstream tasks.

\section{Point-Bind}

The overall pipeline of Point-Bind is shown in Figure~\ref{pipeline}. In Section~\ref{imagebind}, we first provide a preliminary of ImageBind~\cite{girdhar2023imagebind}. Then, in Section~\ref{collection} and~\ref{pointbind}, we elaborate on the training data and multi-modal alignment for Point-Bind, respectively. Finally, in Section~\ref{kkk}, we introduce several 3D-centric applications derived from Point-Bind.

\vspace{0.2cm}
\subsection{Preliminary of ImageBind}
\label{imagebind}
ImageBind~\cite{girdhar2023imagebind} proposes an approach to combine multiple modalities together, which utilizes only image-paired data to learn a joint embedding space of six modalities, i.e., images, text, audio, depth, thermal, and IMU data. It does not need training dataset pairing all six modalities, but leverages the binding property of 2D images, i.e., aligning every single modality to image independently. 
Specifically, ImageBind feeds multi-modal input into corresponding encoders, and adopts 
for cross-modal contrastive learning. 
After training on large-scale image-paired data, ImageBind effectively aligns six modalities into a single shared representation space, enabling emergent cross-modal zero-shot capabilities. 
Based on existing vision-language models, ImageBind can also be utilized for several multi-modal tasks, such as text-to-audio/video retrieval, audio-to-image generation, and audio-referred object detection. Inspired by this, we propose to develop a 3D multi-modal framework that incorporates 3D point clouds with other modalities for general 3D understanding, generation, and instruction following.

\vspace{0.15cm}
\subsection{Training Data}
\label{collection}
To align 3D with multi-modalities, we leverage the pre-trained joint embedding space of ImageBind~\cite{girdhar2023imagebind} and utilize contrastive loss~\cite{zhang2022contrastive, radford2021learning} to simultaneously align 3D point clouds with the other three modalities: image, text, and audio. 
To obtain the contrastive training data, we collect a cross-modal dataset of 3D-image-audio-text pairs. There are three steps for dataset collection as follows.

\paragraph{3D-image-text Pairs.}
We adopt the data pairs of 3D, images, and text from ULIP~\cite{xue2022ulip}, which includes 3D-image-text triplets built from ShapeNet~\cite{chang2015shapenet}, a common-used dataset containing abundant 3D CAD models.
Each 3D point cloud is paired with a corresponding text describing the semantic information of its spatial shape, and a 2D counterpart generated by multi-view image rendering. The text description is constructed by a synset of category names and 64 pre-defined templates.

\vspace{-0.05cm}
\paragraph{3D-audio Pairs.}
To provide more contrastive signals, we also collect the data pairs of 3D and audio from ESC-50~\cite{piczak2015esc} and ShapeNet datasets. Specifically, we first select the categories whose objects can make a sound in the real world from the 55 categories of ShapeNet, such as `airplane', `clock', `washing machine', and `keyboard'. Then, we preserve only the categories that are also within ESC-50.
By this standard, we obtain 9 categories of 3D-audio paired data with extensive audio clips. 

\vspace{-0.05cm}
\paragraph{3D-image-audio-text Pairs Construction.}
Finally, we match each 3D-audio pair with its corresponding 3D-image-text data, resulting in a unified 3D-image-audio-text dataset with extensive cross-modal pairs. During training, we simultaneously feed point clouds and their 
paired data of three modalities for contrastive learning.

\vspace{0.15cm}
\subsection{Aligning 3D with Multi-modality}
\vspace{0.1cm}

\label{pointbind}
After collecting the 3D paired data, we conduct contrastive training to learn a joint embedding space aligning 3D and multi-modalities. Each data sample contains a point cloud $P$, along with the paired 2D image $I$, text description $T^s$, and audio $A$, where $T^s$ represents a set of 64 pre-defined templates.
For the point cloud, we adopt I2P-MAE~\cite{zhang2023learning} as the learnable 3D encoder, denoted as $\operatorname{Encoder_{3D}(\cdot)}$, and append a projection network $\operatorname{Proj(\cdot)}$ of two linear layers, which transforms the encoded 3D feature into ImageBind's multi-modal embedding space. We formulate it as
\begin{align}
\label{3dencode}
\begin{split}
    & F_{3D} = \operatorname{Proj(Encoder_{3D}}(P)),\\
\end{split}
\end{align}
where $F_{3D} \in \mathbb{R}^{1\times C}$ denotes the projected 3D embedding, and $C$ equals the feature dimension of ImageBind. 
For the paired image-text-audio data, we leverage their corresponding encoders from ImageBind for feature extraction, which are frozen during training, formulated as
\begin{align}
\label{encode}
\begin{split}
    F_{2D}, F_{T}^s, F_{A} = \operatorname{ImageBind}(I, T^s, A),
\end{split}
\end{align}
where $F_{2D}, F_{A} \in \mathbb{R}^{1\times C}$ denote the image and audio embeddings, and $F_{T}^s \in \mathbb{R}^{64\times C}$ denotes the text embedding for a set of 64 descriptions. Then, we conduct an average pooling as 
\begin{align}
\label{encode}
\begin{split}
    F_T = \operatorname{Average}(F_{T}^s)\ \ \in \mathbb{R}^{1\times C},
\end{split}
\end{align}
which represents the aggregated text embedding with more robustness.
After that, we adopt contrastive loss~\cite{zhang2022contrastive} between 3D and other modalities, which effectively enforces 3D embeddings to align with the joint representation space, formulated as
\begin{align}
\label{loss}
\nonumber
\begin{split}
    L_{total} = L(F_{3D}, F_{2D}) + L(F_{3D}, F_{T}) + L(F_{3D}, F_{A}).
\end{split}
\end{align}
Note that some categories of the training data do not include the paired audio $A$, since they inherently cannot make any sound, e.g., bottle, planter, and couch, for which we ignore their audio features and loss.

\vspace{0.2cm}
\subsection{Multi-modal Applications}
\label{kkk}

Benefiting from the joint embedding space of Point-Bind, we respectively introduce several emergent application scenarios concerning 3D and multi-modalities.

\paragraph{Any-to-3D Generation.}
Inherited from 2D generative models, existing 3D generation methods can only achieve text-to-3D synthesis. In contrast, with the joint embedding space of Point-Bind, we can generate 3D shapes conditioned on any modalities, i.e., text/image/audio/point-to-mesh. In detail, we directly connect the multi-modal encoders of Point-Bind with the pre-trained decoders of current CLIP-based text-to-3D models, e.g., CLIP-Forge~\cite{sanghi2021clip}. Without further training, we can synthesize a 3D car mesh based on an input car horn.

\paragraph{3D Embedding-space Arithmetic.}
We observe that 3D features encoded by Point-Bind can be directly added with other
modalities to incorporate their semantics, further achieving
composed cross-modal retrieval. 
For instance, the combined embeddings of a 3D car and audio of sea waves can retrieve an image showing a car parking by a beach, while the composition of a 3D laptop and audio of keyboard typing can retrieve an image of someone who is working with a laptop.

\paragraph{3D Zero-shot Understanding.}
For traditional text-inferred 3D zero-shot classification, Point-Bind attains \textit{state-of-the-art} performance guided by additional multi-modal supervision. Besides, Point-Bind can also achieve audio-referred 3D open-world understanding, i.e., recognizing 3D shapes of novel categories indicated by the corresponding audio data~\cite{piczak2015esc}.

\begin{figure*}[t!]
\centering
\includegraphics[width=0.95\textwidth]{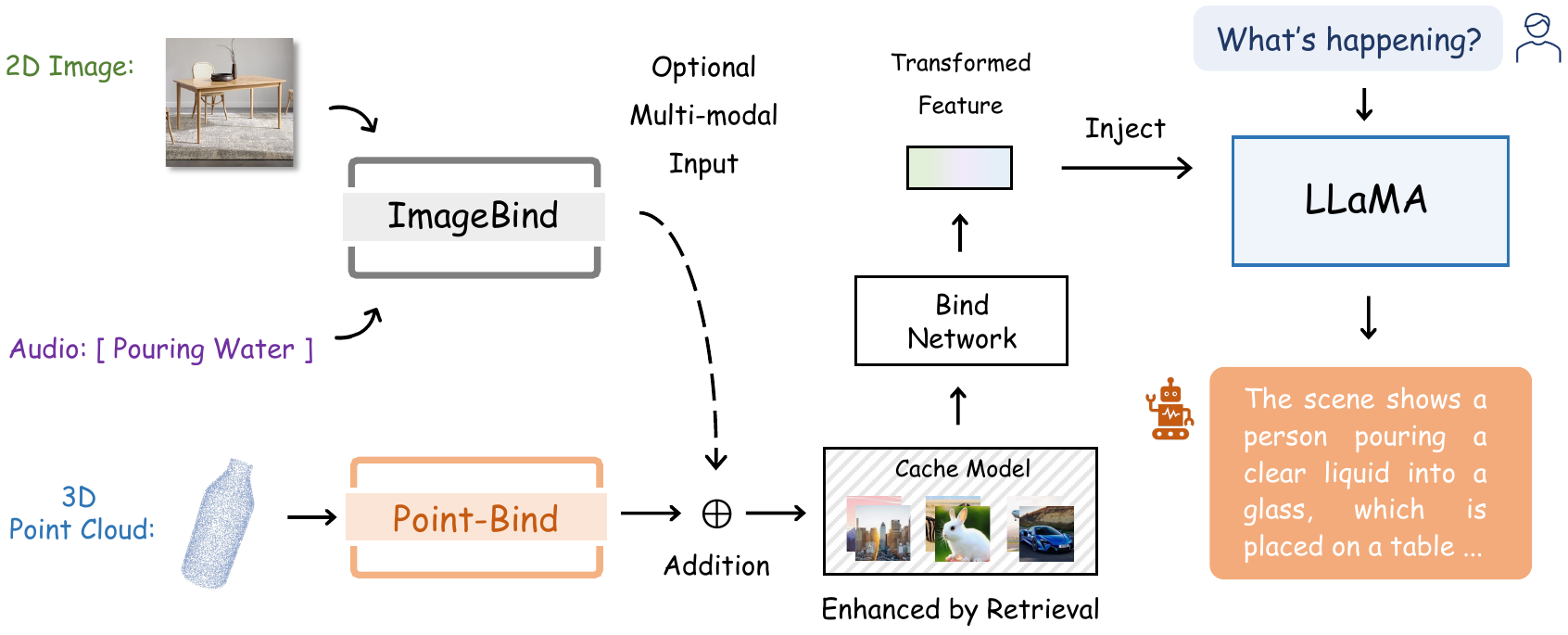}
   \figcaption{\textbf{Inference Paradigm of Point-LLM.} Referring to ImageBind-LLM~\cite{ImageBind-LLM2023}, we adopt a bind network, a visual cache model, and zero-initialized gating mechanisms to fine-tune LLaMA~\cite{touvron2023llama} to follow 3D instructions. Optionally, our Point-LLM can also take as input multi-modality data, and conduct cross-modal reasoning for language response.}
    \label{pointllm}
    \vspace{0.2cm}
\end{figure*}

\vspace{0.2cm}
\section{Point-LLM}
\label{pointllm}
In this section, we illustrate how to leverage Point-Bind to develop 3D large language models (LLMs), termed as Point-LLM, which fine-tunes LLaMA~\cite{touvron2023llama} to achieve 3D question answering and multi-modal reasoning. The overall pipeline of Point-LLM is shown in Figure~\ref{pointllm}.

\vspace{0.2cm}
\subsection{3D Instruction-following Capacity}

Our Point-LLM is developed on top of ImageBind-LLM~\cite{ImageBind-LLM2023}, which conducts multi-modality instruction tuning by injecting the semantics of ImageBind into LLaMA. Our approach exhibits both data and parameter efficiency.

\paragraph{No Need for 3D Instruction Data.} During training, only the public vision-language data~\cite{chen2015microsoft,changpinyo2021conceptual, sharma2018conceptual, schuhmann2022laion} is required for fine-tuning LLaMA to learn the 3D-conditioned response capacity. As Point-Bind has built a joint embedding space between 3D and multi-modalities, if any one of the modalities is trained to connect with LLaMA, the others would also be aligned at the same time. Considering this, we select the 2D image modality, since it has the most public data with paired language. By only aligning ImageBind's image encoder with LLaMA, we avoid the expensive cost of collecting and annotating large-scale 3D instruction data, thus saving extensive resources.

\paragraph{Parameter-efficient Training.}
Instead of tuning the entire LLMs~\cite{zhu2023minigpt, liu2023llava}, we only unfreeze partial parameters within LLaMA for efficient vision-language instruction tuning. Specifically, a learnable bind network is adopted to bridge the image encoder of ImageBind with the language space of LLaMA. Then, a zero-initialized gating mechanism is proposed to add the image features after the bind network to the words tokens within LLaMA. This mechanism can progressively inject visual instruction cues into LLaMA for stable training at early stages, inspired by LLaMA-Adapter~\cite{zhang2023llama}. By such a parameter-efficient fine-tuning strategy, most parameters of LLaMA are kept frozen, and only the zero-initialized gating factors and bias-norm weights~\cite{zhang2023llama} are learnable for fine-tuning. Please refer to ImageBind-LLM~\cite{ImageBind-LLM2023} for further training details. After the vision-language training, the joint embedding space enables LLaMA to naturally align with other modalities, such as audio within ImageBind and 3D point clouds of Point-Bind. Therefore, our Point-LLM effectively provides LLaMA with 3D-instruction following capacity without any 3D instruction data, indicating superior data efficiency.

\vspace{0.15cm}
\subsection{3D Question Answering}
For an input language instruction and a 3D point cloud, we feed them into the fine-tuned LLaMA and our Point-Bind, respectively. Then, the encoded 3D feature is enhanced by a visual cache model proposed in ImageBind-LLM, before feeding into the bind network. The cache model is only adopted during inference, and constructed in a training-free manner~\cite{zhang2022tip}.

\vspace{-0.1cm}
\paragraph{Enhancement by Visual Cache.} As we adopt the image encoder of ImageBind for training, but switch to Point-Bind's 3D encoder for inference, the cache model is designed to alleviate such modality discrepancy for better 3D geometry understanding. Referring to ImageBind-LLM, the cache model stores from three ImageBind-encoded image features from the training data, which are regarded as both keys and values for knowledge retrieval.
We regard the input 3D feature as the query, and retrieve the top-$k$ similar visual keys from the cache model. Then, according to the cosine similarity, we aggregate the corresponding cached values (top-$k$ similar image features), and add the result to the original 3D feature via a residual connection. In this way, the enhanced 3D feature can adaptively incorporate similar visual semantics from the cache model. This boosts the representation quality of 3D shapes, and mitigates the semantic gap of 2D-3D encoders within Point-LLM. After this, the enhanced feature is fed into the bind network for feature transformation and LLaMA for response generation.

\paragraph{3D and Multi-modal Reasoning.}
In addition to point clouds, our Point-LLM can also conduct cross-modal reasoning and generate responses conditioned on multiple modalities. For an additional input image or audio, we utilize the image or audio encoder of ImageBind to extract the features, and directly add them with the 3D feature encoded by Point-Bind. By injecting such integrated features into LLaMA, Point-LLM can reason cross-modal semantics, and respond with the information of all input modalities. This demonstrates the promising significance of aligning multi-modality with 3D LLMs.

\begin{figure*}[t!]
\centering
\includegraphics[width=\textwidth]{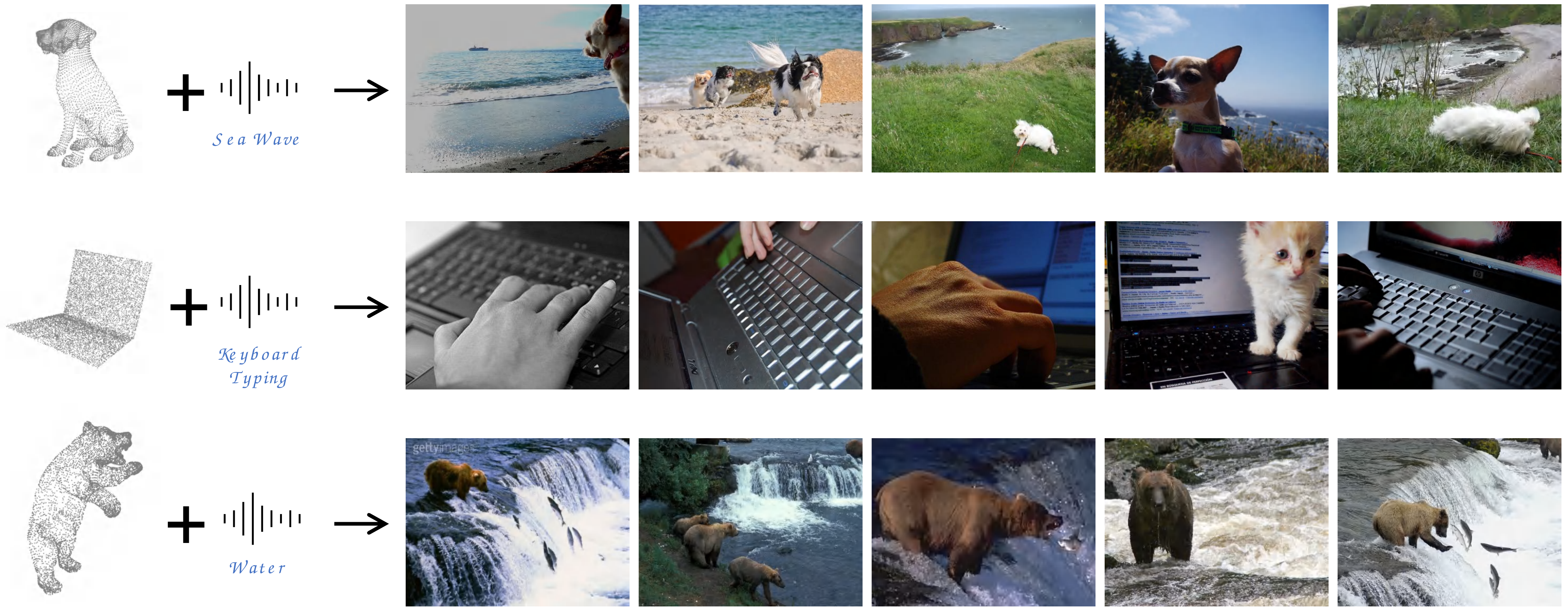}
   \figcaption{\textbf{Embedding-space Arithmetic of 3D and Audio.} We demonstrate Point-Bind's capability for multi-modal semantic composition by retrieving 2D images with a combination of 3D point cloud and audio embeddings.}
    \label{compose}
\end{figure*}

\section{Experiments}
In this section, we first present the implementation details of the multi-modal training for Point-Bind. Then, we illustrate the emergent multi-modal applications, i.e., Point-LLM for 3D instruction following, 3D cross-modal retrieval, 3D embedding-space arithmetic, any-to-3D generation, and 3D zero-shot understanding. Finally, we conduct an ablation study to verify the effectiveness of our designs.

\subsection{Implementation Details}
We adopt pre-trained I2P-MAE~\cite{zhang2023learning} as the 3D encoder of Point-Bind, and utilize the collected 3D-image-text-audio pairs for pre-training. We only update the 3D encoder with the newly added projection network, and freeze the encoders of other modalities in ImageBind~\cite{girdhar2023imagebind}. The projection network is simply composed of two linear layers with an intermediate LayerNorm~\cite{ba2016layer}.
We train Point-Bind for 300 epochs with a batch size of 64, and adopt AdamW~\cite{loshchilov2017decoupled} as the optimizer with a learning rate of 0.003. 

\subsection{Point-LLM for 3D Q\&A}
\label{3dllm_eval}
\paragraph{Settings.}
We refer to ImageBind-LLM~\cite{ImageBind-LLM2023} to conduct parameter- and data-efficient fine-tuning to inject 3D instructions into the pre-trained LLaMA 7B model~\cite{touvron2023llama}. In detail, the fine-tuning techniques include zero-initialized gating~\cite{gao2023llama,zhang2023llama}, LoRA~\cite{hu2021lora}, and bias-norm tuning~\cite{xie2023difffit,zaken2021bitfit,frankle2020training,giannou2023expressive}. We utilize a collection of several datasets~\cite{changpinyo2021conceptual, sharma2018conceptual, schuhmann2022laion} for vision-language training, and require no 3D instruction-following dataset due to the learned joint embedding space.

\paragraph{Analysis.}
In Figure~\ref{3dllm}, we provide the question-answering examples of Point-LLM, which shows favorable 3D instruction-following and multi-modal reasoning capacity. As shown, for either English or Chinese instructions, Point-LLM can effectively incorporate the spatial geometry of input point clouds and generate detailed language responses.
It obtains a comprehensive 3D understanding for both global and local characteristics, e.g., recognizing the pattern of the piano keyboard and the shape of the airplane's wing and tail.
Then, our Point-LLM can also respond with cross-modal understanding. For an input 3D model with a 2D image or audio, Point-LLM can enable LLaMA to take both two conditions into understanding and reasoning, which thus incorporates multi-modal semantics in the output language response. With superior data- and parameter-efficiency, the examples indicate the 3D multi-modal instruction-following capabilities of Point-LLM.

\begin{table}[t]
\small
\centering
\tabcaption{\textbf{Performance on 3D Cross-modal Retrieval}, including 3D-to-3D, 2D-to-3D, 3D-to-2D, and text-to-3D retrieval. We report the mAP scores (\%) on ModelNet40~\cite{wu20153d} dataset.}
\label{retrieve}
\begin{adjustbox}{width=\linewidth}
	\begin{tabular}{lcccc}
	\toprule
		\makecell*[l]{Method} &3D\ $\rightarrow$\ 3D &2D\ $\rightarrow$\ 3D  &3D\ $\rightarrow$\ 2D &Text\ $\rightarrow$\ 3D \\
		\cmidrule(lr){1-1} \cmidrule(lr){2-2} \cmidrule(lr){3-3} \cmidrule(lr){4-4} \cmidrule(lr){5-5}
            PointCLIP~\cite{zhang2022pointclip} &37.63 &13.12 &5.28 &10.86 \\
            PointCLIP-V2~\cite{Zhu2022PointCLIPV2} &47.94 &20.48 &9.22 &52.73\\
            ULIP~\cite{xue2022ulip} &60.58 &20.30 &29.75 &50.51 \\
            \rowcolor{gray!8} \textbf{Point-Bind} &\textbf{63.23} &\textbf{34.59} &\textbf{42.83} &\textbf{64.50} \\            
	     \textit{Gain} &\textcolor{citecolor}{+2.65} &\textcolor{citecolor}{+14.29} &\textcolor{citecolor}{+13.08} &\textcolor{citecolor}{+13.99}\\
	  \bottomrule
	\end{tabular}
\end{adjustbox}
\end{table}


\subsection{3D Cross-modal Retrieval}
To evaluate the multi-modal alignment of Point-Bind, we experiment on several cross-modal retrieval tasks, i.e., 3D-to-3D, 2D-to-3D, 3D-to-2D, and text-to-3D retrieval. 

\begin{figure*}[t!]
\centering
\includegraphics[width=\textwidth]{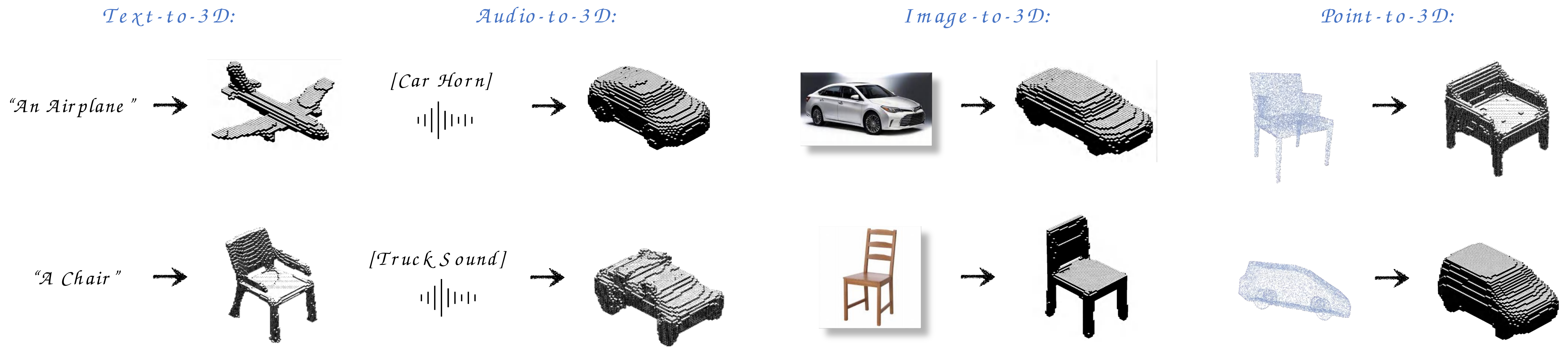}
   \figcaption{\textbf{Any-to-3D Generation.} Besed on CLIP-Forge~\cite{sanghi2021clip}, our constructed joint embedding space can effectively generate 3D mesh models conditioned on text, audio, image, and point cloud input.}
    \label{3dgen}
\vspace{0.1cm}
\end{figure*}

\begin{table}[t]
\vspace{0.05cm}
\small
\centering
\tabcaption{\textbf{Performance of 3D Zero-shot Classification.} We report the classification accuracy (\%) on ModelNet40~\cite{wu20153d}.}
\label{zs}
	\begin{tabular}{lcc}
	\toprule
		\makecell*[l]{Model} &Encoder &Performance\\
		\cmidrule(lr){1-1} \cmidrule(lr){2-2} \cmidrule(lr){3-3} 
            PointCLIP~\cite{zhang2022pointclip} &CLIP &20.2 \\
            ULIP~\cite{xue2022ulip} &Point-BERT &60.4 \\
            PointCLIP V2~\cite{Zhu2022PointCLIPV2} &CLIP &64.2 \\
            ULIP 2~\cite{xue2023ulip2} &Point-BERT &66.4 \\
            \rowcolor{gray!8} Point-Bind &Point-BERT &76.3 \\            
            \rowcolor{gray!8} \textbf{Point-Bind} &\textbf{I2P-MAE} &\textbf{78.0} \\
	     \textit{Gain} &- &\textcolor{citecolor}{+11.6}\\
	  \bottomrule
	\end{tabular}
\vspace{0.2cm}
\end{table}

\paragraph{Settings.}
We conduct 3D zero-shot retrieval on multi-modal ModelNet40~\cite{wu20153d} dataset, which contains 9,843 CAD models for training and 2,468 for testing of 40 categories. ModelNet40 provides data of three modalities for retrieval, i.e., image, point cloud, and mesh. We obtain the retrieved results by ranking the similarities between embeddings of point clouds and other modalities. 
Following previous works~\cite{jing2021cross, liu2023instance}, we measure the networks via the mean Average Precision (mAP) score, a commonly used evaluation criterion for retrieval tasks. 

\vspace{-0.1cm}
\paragraph{Analysis.}
In Table~\ref{retrieve}, we report the quantitive results for 3D zero-shot retrieval, where Point-Bind attains \textit{state-of-the-art} performance on all benchmarks compared with prior works. In particular, for 2D-to-3D and text-to-3D retrieval, Point-Bind surpasses the second-top ULIP~\cite{xue2022ulip} significantly by \textbf{+14.29\%} and \textbf{+13.99\%} improvements, respectively. This indicates the superior cross-modal understanding capacity of our approach.

\vspace{0.2cm}
\subsection{Embedding-space Arithmetic with 3D}
With the multi-modal alignment, we further explore the capability of embedding composition, i.e., the embedding-space arithmetic of 3D and other modalities, e.g., audio. 

\paragraph{Settings.} 
To obtain the multi-modal input for arithmetic, we utilize 3D objects from ShapeNet~\cite{chang2015shapenet} and TextANIMAR2023~\cite{animal}, and audio clips from ESC-50~\cite{piczak2015esc}. 
We simply add the 3D and audio embeddings from Point-Bind and ImageBind, respectively, and then retrieve 2D images from ImageNet~\cite{deng2009imagenet} with 1,000 image categories. 

\paragraph{Analysis.}
In Figure~\ref{compose}, we show the results of 2D image retrieval with the composed embeddings between 3D and audio. As shown in the first row, with the combined embeddings of a 3D dog and sea-wave audio, we effectively retrieve 2D images of dogs by the sea. Similarly, with the combination of a 3D laptop and keyboard-typing audio, the obtained images show someone is working with a laptop, or a cat inadvertently presses on the keyboard. Likewise, the last row retrieves images of bears hunting by the water by using embeddings of a 3D bear and audio of flowing water. The examples demonstrate that the 3D features encoded by Point-Bind can be directly added with other modalities, and well incorporate their semantics, achieving favorable composed cross-modal retrieval capacity.

\vspace{0.25cm}
\subsection{Any-to-3D Generation}
\paragraph{Settings.}
Existing text-to-3D generation methods normally adopt CLIP's text encoder to process the input language prompt. Considering this, we simply replace it with the multi-modalities encoders of Point-Bind and ImageBind without further training, which follows the original generative decoder for 3D shape synthesis. We adopt the decoder of CLIP-Forge~\cite{sanghi2021clip} by default.

\paragraph{Analysis.} In Figure~\ref{3dgen}, we show the examples of any-to-3D generation powered by Point-Bind. For text, audio, and point cloud prompts, our approach can all produce satisfactory 3D meshes. This demonstrates the well-aligned embedding space of 3D and multiple modalities.

\vspace{0.25cm}
\subsection{3D Zero-shot Understanding}
In this section, we test the open-word understanding ability of Point-Bind, i.e., recognizing novel classes, by 3D zero-shot classification on ModelNet40~\cite{wu20153d} dataset.

\paragraph{Settings.}
Following previous works, we utilize the text embeddings from CLIP's~\cite{radford2021learning} or ImageBind~\cite{girdhar2023imagebind}'s text encoder to construct the zero-shot classification head. Specifically, we apply a simple template of \textit{`a/an [CLASS]'} for the 40 categories of ModelNet40, and calculate the cosine similarity between 3D and all textual embeddings, selecting the most similar one as the final prediction.

\paragraph{Analysis.}
We report the 3D zero-shot classification accuracy in Table~\ref{zs}, where our Point-Bind surpasses existing methods with \textit{state-of-the-art} performance. This indicates the unified representation space of Point-Bind leads to strong emergent 3D open-world recognition.

\begin{table}[t!]
\centering
\small
\tabcaption{\textbf{Ablation Study} exploring different designs of the projection network and 3D encoders. We report the results (\%) for zero-shot classification on ModelNet40~\cite{wu20153d}.}
\label{abla}
	\begin{tabular}{lc|lc}
	\toprule
 Projection &Acc. &3D Encoder &Acc.  \\
  \cmidrule(lr){1-1} \cmidrule(lr){2-2} \cmidrule(lr){3-3} \cmidrule(lr){4-4} 
  One Linear &76.46  &PointNeXt~\cite{qian2022pointnext} &67.96 \\ 
  Two Linear &\textbf{78.00} &Point-BERT~\cite{yu2021pointbert} &76.70 \\ 
  Three Linear &76.78 &I2P-MAE~\cite{zhang2023learning}   &\textbf{78.00} \\ 
	  \bottomrule
	\end{tabular}
\vspace{0.1cm}
\end{table}

\subsection{Ablation Study}
To investigate the effectiveness of our designs in Point-Bind, we conduct ablation studies on the projection network and 3D encoders in Table~\ref{abla}. We report the performance of zero-shot classification on ModelNet40~\cite{wu20153d} dataset.
In the first two columns, we experiment with different projection schemes for embeddings after the 3D encoder. As shown, using two linear layers for embedding projection performs the best.
In the last two columns, we utilize different 3D encoders in Point-Bind, i.e., Point-BERT~\cite{yu2021pointbert}, PointNeXt~\cite{qian2022pointnext}, and I2P-MAE~\cite{zhang2023learning}. As reported, the self-supervised Point-BERT and I2P-MAE achieve much better performance, indicating the importance of 3D pre-training to boost the multi-modal alignment.

\section{Conclusion}
In this paper, we propose \textbf{Point-Bind}, a 3D multi-modality model that aligns 3D point clouds with multi-modalities, guided by ImageBind. 
By aligning 3D objects with their corresponding image-audio-text pairs, Point-Bind obtains a joint embedding space, and exhibits promising 3D multi-modal tasks, such as any-to-3D generation, 3D embedding arithmetic, and 3D open-world understanding.
Upon that, we further introduce \textbf{Point-LLM}, the first 3D large language model (LLM) with instruction-following capability in both English and Chinese.
Our future work will focus on aligning multi-modality with more diverse 3D data, such as indoor and outdoor scenes, which allows for wider application scenarios.

{\small
\bibliographystyle{ieee_fullname}
\bibliography{egbib}
}

\end{document}